\documentclass[10pt,twocolumn,letterpaper]{article}

\usepackage[svgnames]{xcolor}
\usepackage{cvpr}
\usepackage{times}
\usepackage{epsfig}
\usepackage{epstopdf}
\usepackage{graphicx}
\usepackage{amsmath}
\usepackage{amssymb}
\usepackage{textcomp}
\usepackage{gensymb}
\usepackage{booktabs}
\usepackage{bbm}
\usepackage{bm}
\usepackage[nocompress]{cite}
\usepackage{changepage}
\usepackage{url}
\providecommand\cc[1]{\textcolor{Black}{#1}}
\providecommand\changes[1]{\textcolor{Black}{#1}}
\providecommand\JD[1]{\textcolor{Black}{#1}}
\providecommand\JDtwo[1]{\textcolor{Black}{#1}}
\providecommand\KG[1]{\textcolor{Black}{#1}}

\cvprfinalcopy 

\def\cvprPaperID{1598} 

\ifcvprfinal\fi
\begin{document}

\title{Learning to Look Around:\\Intelligently Exploring Unseen Environments for Unknown Tasks}

\author{Dinesh Jayaraman\\
UC Berkeley\\
{\tt\small dineshjayaraman@berkeley.edu}\thanks{work done while author was a PhD student at UT Austin}
\and
Kristen Grauman\\
UT Austin\\
{\tt\small grauman@cs.utexas.edu}
}

\maketitle
\vspace*{-0.2in}
\begin{abstract}
  \vspace{-0.05in}

It is common to implicitly assume access to intelligently captured inputs (e.g., photos from a human photographer), yet autonomously capturing good observations is itself a major challenge.  We address the problem of \emph{learning to look around}: 
if a visual agent has the ability to voluntarily acquire new views to observe its environment, how can it learn efficient exploratory behaviors to acquire informative observations?
We propose a reinforcement learning solution, where the agent is rewarded for actions that reduce its uncertainty about the unobserved portions of its environment. Based on this principle, we develop a recurrent neural network-based approach to perform \emph{active completion} of panoramic natural scenes and 3D object shapes. Crucially, the learned policies are not tied to any recognition task nor to the particular semantic content seen during training.  As a result, 1) the learned ``look around" behavior is relevant even for new tasks in unseen environments, and 2) training data acquisition involves no manual labeling. Through tests in diverse settings, we demonstrate that our approach learns useful \emph{generic} 
policies that transfer to new unseen tasks and environments. Exploration episodes are shown at \url{https://goo.gl/BgWX3W}.
\end{abstract}

\vspace{-0.1in}
\section{Introduction}
\vspace{-0.05in}
Visual perception requires not only making inferences from observations, but also making decisions about \emph{what to observe}. Individual views of an environment afford only a small fraction of all information relevant to a visual agent. For instance, an agent with a view of a television screen in front of it may not know if it is in a living room or a bedroom. An agent observing a mug from the side may have to move to see it from above \JD{to know what is inside}.  {An agent surveying a rescue site may need to explore at the onset to get its bearings.}

In principle, complete certainty in perception is only achieved by making every possible observation---that is, looking around in all directions, or systematically examining all sides of an object---yet observing all aspects is often inconvenient if not intractable. In practice, however, not all views are equally informative.  The natural visual world contains regularities, suggesting not every view needs to be sampled for near-perfect perception. For instance, humans rarely need to fully observe an object to understand its 3D shape~\cite{soska2008development,soska2010systems,kellman1983perception}, and one can often understand the primary contents of a room without literally scanning it~\cite{torralba2006contextual}. Given a set of past observations, some new views are more useful than others. This leads us to investigate the question: \emph{how can a learning system make intelligent decisions about how to acquire new exploratory visual observations?}

Today, much of the computer vision literature deals with inferring visual properties {from a fixed observation}. For instance, there are methods to infer shape from multiple views~\cite{hartley2003multiple}, depth from monocular views~\cite{saxena2009make3d}, or category labels of objects~\cite{krizhevsky2012imagenet}. The implicit assumption is that the input visual observation is already appropriately captured. We contend that this assumption neglects a key part of the challenge: intelligence is often required to obtain proper inputs in the first place. {Arbitrarily} framed snapshots of the visual world are ill-suited both for human perception~\cite{edelman1992orientation,palmer1981canonical} and for machine perception~\cite{ammirato2017dataset,snap-points}. Circumventing the acquisition problem is only viable for passive perception algorithms running on disembodied stationary machines, which are tasked only with processing human-captured imagery.

\begin{figure*}[t]
  \centering
  \includegraphics[width=1\linewidth]{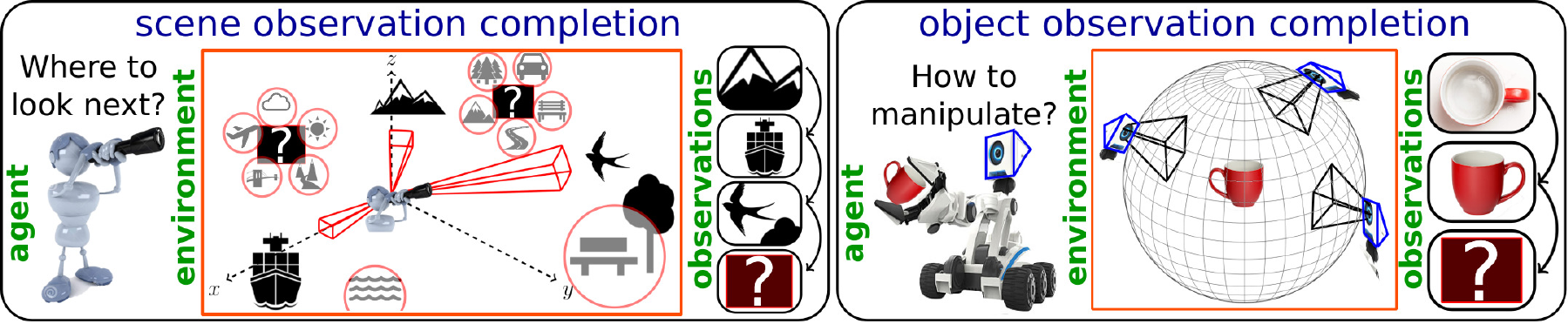}
  \caption{\small{Looking around efficiently is a complex task requiring the ability to reason about regularities in the visual world using cues like context and geometry. (Left) An agent that has observed limited portions of its environment can reasonably hallucinate some unobserved portions (e.g. water near the ship), but is much more uncertain about other portions. Where should it look next? (Right) An agent inspecting a mug. Having seen a top view and a side view, how must it rotate the mug now to get maximum new information? Critically, we aim to learn policies that are not specific to a given object or scene, and not even to a specific individual task. Rather, the look-around policies ought to benefit the agent exploring new, unseen environments and performing tasks unspecified when learning the look-around behavior.}
  }
  \label{fig:conceptfig}
  \vspace{-0.13in}
\end{figure*}

In contrast, we are interested in \emph{learning to observe} efficiently---a critical yet understudied problem for autonomous embodied visual agents. An agent ought to be able to enter a new environment or pick up a new object and intelligently (non-exhaustively) look around.  This capability would be valuable in both task-driven scenarios (e.g., a drone searches for signs of a particular activity) as well as scenarios where the task itself unfolds simultaneously with the agent's exploratory actions (e.g., a search-and-rescue robot enters a burning building and dynamically decides its mission).  While there is interesting recent headway in active object recognition~\cite{dinesh-eccv2016,GERMS,shapenet,ammirato2017dataset} and intelligent search mechanisms for detection~\cite{yeung2016end,lazebnik-iccv2015,mathe-cvpr2016,timely-nips2012}, such systems are supervised and task-specific---limited to accelerating a pre-defined recognition task.


We address the general setting, where exploration is not specialized to one task, but should benefit perception tasks in general. To this end, we formulate an unsupervised learning objective based on \emph{active observation completion}: a system must intelligently acquire a small set of observations from which it can hallucinate all other possible observations.  The agent continuously updates its internal model of a target scene or 3D shape based on all previously observed views.  The goal is not to produce photorealistic predictions, but rather to represent the agent's evolving internal state.  Its task is to select actions leading to new views {that will} efficiently complete its internal model. Posing the active view acquisition problem in terms of observation completion has two key advantages{:  generality and low cost (label-free) training data.} It is also well-motivated by findings that infants' abilities to actively manipulate and inspect objects correlates with learning to complete 3D shapes~\cite{soska2010systems}.

We develop a reinforcement learning solution for active visual completion.  Our approach uses recurrent neural networks to aggregate information over a sequence of views. The agent is rewarded based on its predictions of unobserved views.



\JDtwo{We explore our idea in two settings.} See Figure~\ref{fig:conceptfig}. In the first, the agent scans a scene through its limited field of view camera; the goal is to select efficient camera motions so that after a few glimpses, it can model unobserved portions of the scene well. In the second, the agent manipulates a 3D object to inspect it; the goal is to select efficient manipulations so that after only a small number of actions, it has a full model of the object's 3D shape. In both cases, the system must learn to leverage visual regularities (shape primitives, context, etc.) that suggest the likely contents of unseen views, focusing on portions that are hard to hallucinate. Furthermore, we show our exploratory policies are generic enough to be \emph{transferred} to entirely new unseen tasks and environments.







\vspace*{-0.05in}
\section{Related work}~\label{sec:related}
\vspace*{-0.1in}

\paragraph{Saliency and attention:} Previous work studies the question of ``where to look'' to prioritize portions of \emph{already captured} image/video data,
so as to reserve computation for the most salient regions or block out distractors
~\cite{harel2006graph,torralba-mask,liu2011learning,achanta2009frequency,perazzi2012saliency,Mnih2014-dg,Ba2014-rr,Xu2015-vc,Bazzani2011-cb}\changes{, or to predict the gaze or preference of a human observer~\cite{li2013learning,deep-pilot,su2016pano2vid}.} 
In contrast, in our setting, the system can never observe a snapshot of its entire environment at once; {its decision is not where to focus within a current observation, but rather where to look for a \emph{new} observation.} \cc{We compare to a saliency-based method.}

\vspace*{-0.15in}
\paragraph{Optimal sensor placement:} 
{The sensor placement literature studies} how to place sensors in a distributed network to provide maximum coverage~\cite{dhillon2003sensor,krause2007near,wang2011coverage}.  
Unlike our active completion problem, the sensors are static, i.e., their positions are preset, and their number is fixed. Further, sensor placement is based on coverage properties of the sensors,
whereas our model must \emph{react} to past observations.

\vspace*{-0.15in}
\paragraph{Active perception:} Intelligent  control strategies for visual tasks 
were pioneered by~\cite{bajcsy1988active,aloimonos1988active,ballard1991animate,wilkes1992active}. 
Recent work considers tasks such as active object localization~\cite{lazebnik-iccv2015,timely-nips2012,
mathe-cvpr2016,Andreopoulos2013-bm,gonzalez2014active,Soatto2009-pk,gupta2017cognitive,zhu2016target,mirowski2016learning}, action detection in video~\cite{yeung2016end}, and object recognition\KG{~\cite{GERMS,JohnsCVPR2016,dinesh-eccv2016,ammirato2017dataset}}
{including foveated vision systems that selectively obtain higher resolution data~\cite{ng-foveated,butko-movellan,ranzato2014learning}}.

Our idea stands out from this body of work in two key aspects: (1) Rather than target a pre-defined recognition task, we aim to learn a data acquisition strategy useful to perception in general, hence framing it as active ``observation completion''. \changes{We show how policies trained on our task are useful for recognition tasks \emph{for which the system has not been trained to optimize its look-around behavior}.}
(2) {Rather than manually labeled data, our method learns from unlabeled observations.} Training good policies usually requires large amounts of data; our unsupervised objective removes the substantial burden of manually labeling this data. \changes{Instead, our approach exploits viewpoint-calibrated observations as ``free'' annotations that an agent can acquire through its own explorations at training time.}

Work on intrinsic motivation ``pseudorewards''~\cite{machado} also reduces the need for external supervision, but \JDtwo{focuses on learning ``options''} for policies seeking reward signals in a specific task and fixed environment. Similarly motivated self-supervised work~\cite{pathak-curiosity}  
\KG{learns policies to play sparse-reward video games by \JDtwo{augmenting environmental reward from the game engine with rewards for actions whose outcomes are unpredictable}.} \KG{Neither work explores problems with real natural images.}




\vspace*{-0.15in}
\paragraph{Active visual localization and mapping:} Active visual SLAM aims to limit samples needed to densely reconstruct a 3D environment using geometric methods~\cite{martinez2007active,davison2002simultaneous,kollar2008trajectory,kim2013perception,spica2014active}.
Beyond measuring uncertainty in the current scene, our learning approach capitalizes on learned context from previous experiences with \emph{different} scenes/objects.  \cc{While purely geometric methods are confined to using exactly what they see, and hence typically require dense observations, our approach can infer substantial missing content \JD{using semantic and contextual cues.}}

\vspace*{-0.15in}
\paragraph{Image completion:}  \cc{Completion tasks appear in other contexts within vision and graphics.} Inpainting and texture synthesis fill small holes
(e.g.,~\cite{pathak2016context,texture-synthesis-bethge}), and large holes can be filled by pasting in regions from similar-looking scenes~\cite{hays2007scene}.  Recent work explores unsupervised ``proxy tasks" to learn representations, via various forms of completion like inpainting and colorization~\cite{pathak2016context,zhang2016split,larsson2017colorization}.
Our observation completion setting differs from these in that {1)} it requires agent \emph{action}, {2)} a much smaller fraction of the overall environment is observable at a time, 3) \JD{our target is a representation of multimodal beliefs, rather than a photorealistic rendering}, and {4)} we use completion to learn exploratory \emph{behaviors} {rather than features}.

\vspace*{-0.15in}
\paragraph{{Learning to reconstruct:}} While 3D vision has long been tackled with geometry and densely sampled views~\cite{hartley2003multiple}, recent work explores ways to inject learning into reconstruction and view synthesis~\cite{KulkarniGraphicsNet2015,ChairsCVPR2015,ZhouViewSynthesis2017,
ChoySavarese2016,Freeman-ECCV2016,YanPerspective2016,hane2017hierarchical,my_1shot}. {Whereas prior work learns to aggregate and extrapolate from passively captured views in one shot, our work is the first to consider active, sequential acquisition of informative views.  
  Our view synthesis module builds on the one-shot reconstruction approach of~\cite{my_1shot}, but our contribution is entirely different. Whereas~\cite{my_1shot}  \emph{infers a viewgrid image} from a single input view, our approach \emph{learns look-around behavior} to select the sequence of views expected to best reconstruct all views. \cc{Further, while~\cite{my_1shot} targets image feature learning, we aim to learn exploratory \emph{action policies}.}

\section{Approach}~\label{sec:approach}
\vspace*{-0.05in}

We now present our approach for learning to actively look around.  For ease of presentation, we present the problem setup as applied to a 3D object understanding task.  With minor modifications (detailed in Sec.~\ref{sec:exp}) our framework applies also to the panoramic scene understanding setting.  Both will be tested in results.

\subsection{Problem setup and notation}


The problem setting is as follows: At timestep $t=1$, an active agent is presented with an object $X$ in a random, unknown pose\footnote{We assume the elevation angle alone is known, since this is true of real-world settings due to gravity.}. At every timestep, it can perform one action to rotate the object and observe it from the resulting viewpoint. Its objective is to make efficient exploratory rotations to understand the object's shape. It maintains an internal representation of the object shape, which it updates after every new observation. After {a budget of} $T$ timesteps of exploration, it \JD{should have learned a model that can} produce a view of the object \JD{as seen from any specified new viewing angle.}

We discretize the space of all viewpoints into a ``viewgrid'' $V(X)$. To do this, we evenly sample $M$ azimuths from 0\degree~to 360\degree~and $N$ elevations from -90\degree~to +90\degree~and form all $MN$ possible {pairings}. Each pairing of an azimuth and an elevation corresponds to one viewpoint $\bm{\theta}_i$ on a viewing sphere focused on the object. Let $\bm{x}(X,\bm{\theta}_i)$ denote the 2D image corresponding to the view of object $X$ from viewpoint $\bm{\theta}_i$. {The} viewgrid $V(X)$ is the table of views $\bm{x}(X,\bm{\theta_i})$ for $1\leq i \leq MN$. During training, the full viewgrid of each object is available to the agent as supervision. During testing, the system must predict {the} complete viewgrid, having seen only a few views within it.

At each timestep $t$, the agent observes a new view $\bm{x}_t$ and updates its \emph{prediction} for the viewgrid ${\hat{V}_t(\bm{x_1},\cdots,\bm{x_t})}$. Simplifying notation, the problem now reduces to sequentially exploring the viewgrid $V$ to improve ${\hat{V}_t}$ --- in other words, actively \emph{completing the observation} of the viewgrid $V(X)$ of object $X$.  Given the time budget $T<<MN$, the agent can see a maximum of $T$ views out of all $MN$ views (maximum because it is allowed to revisit old views).

We explicitly choose to complete the viewgrid in the pixel-space so as to maintain generality---the full scene/3D object encompasses all potentially useful information for \emph{any} task. Hence, by formulating active observation completion in the pixel space, our approach avoids committing to any intermediate semantic representation, in favor of learning policies that seek generic information useful to many tasks. That said, our formulation is easily adaptable to more specialized settings---e.g., if the target task only requires perceiving poses of people, the predictions could be in the keypoint space instead.

The active observation completion task poses three major challenges.  Firstly, to predict unobserved views well, the agent must learn to understand 3D from very few views.  Classic geometric solutions 
struggle under these conditions. Instead, reconstruction must draw on semantic and contextual cues. 
  Secondly, intelligent action is critical to this task.  Given a set of past observations, the system must act based on {which} new views are likely to be most informative, i.e., determine which views would most improve its model of the full viewgrid. We stress that the system will be faced with objects and scenes it has never encountered during training, yet still must intelligently choose where it would be valuable to look next.  \cc{Finally, the task is highly underconstrained---after only a few observations, there are typically  many possibilities, and the agent must be able to handle this multimodality.}  




\begin{figure*}[t]
  \centering
  \includegraphics[width=0.9\linewidth]{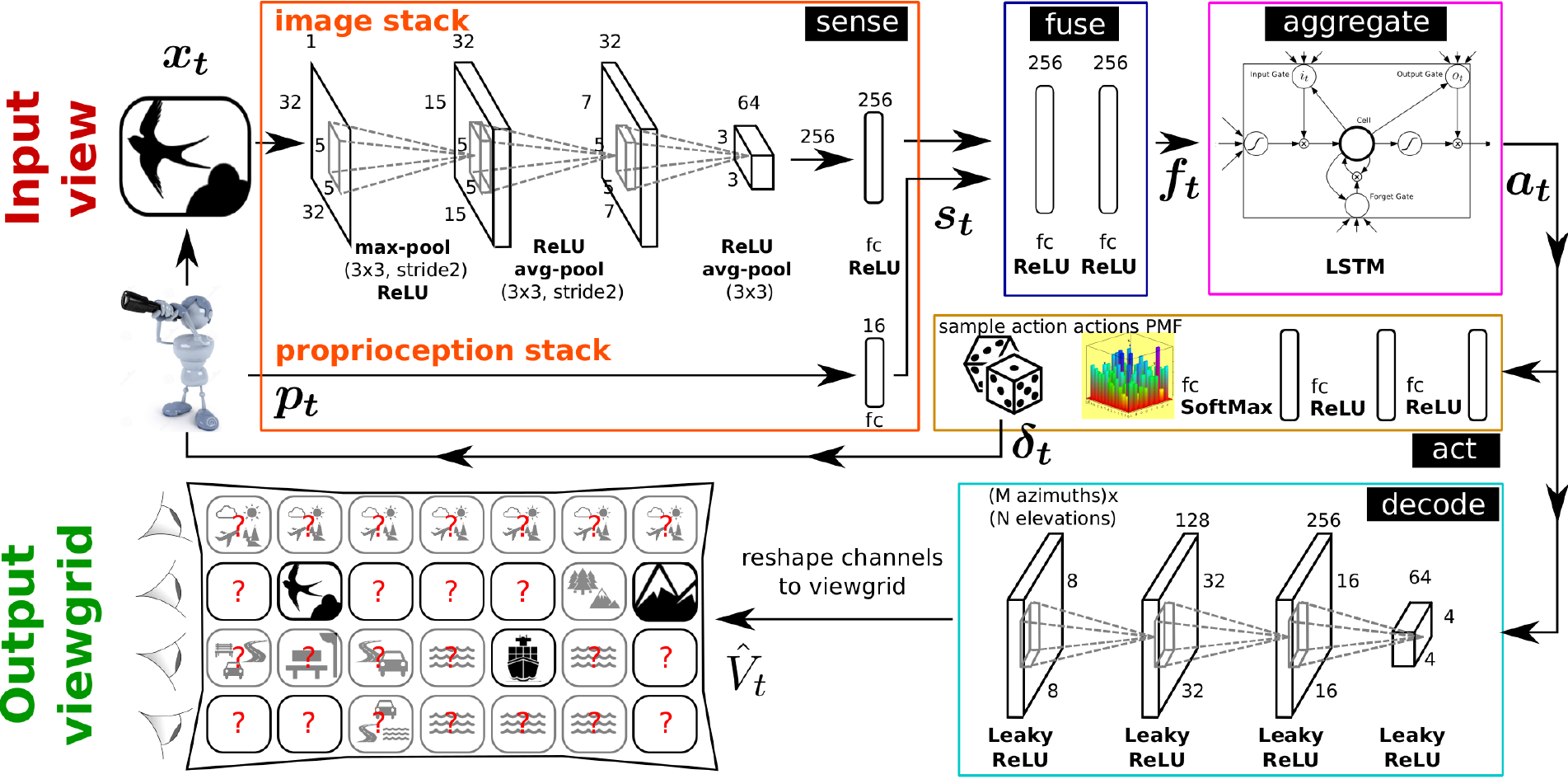} 
  \caption{Architecture of our {active observation completion} system. {While the input-output pair shown here is for the case of $360\degree$ scenes, we use the same architecture for \KG{the case of 3D objects}}. \JDtwo{In the output viewgrid, solid black portions denote observed views, question marks denote unobserved views, and transparent black portions denote the system's uncertain contextual guesses.} See Sec.~\ref{sec:approach-core} for details.}
  \label{fig:architecture}
  \vspace{-0.13in}
\end{figure*}
\subsection{Active observation completion framework}~\label{sec:approach-core}
\vspace{-0.15in}

Our solution to these challenges is a recurrent neural network, whose architecture naturally splits into five modules with distinct functions: \textsc{sense}, \textsc{fuse}, \textsc{aggregate}, \textsc{decode}, and \textsc{act}.   {We first present these modules and their connections; Sec.~\ref{sec:objective} below defines the learning objective and optimization.}  Architecture details for all modules are given in Fig~\ref{fig:architecture}.

\vspace*{-0.1in}
\paragraph{Encoding to an internal model of the target} {First we define the core modules with which the agent encodes its internal model of the current environment.  }At each step $t$, the agent is presented with a 2D view $\bm{x}_t$ captured from a new viewpoint $\bm{\theta}_t$. {We stress that absolute viewpoint coordinates $\bm{\theta}_t$ are \emph{not} fully known, and objects/scenes are \emph{not} presented in any canonical orientation. {All viewgrids inferred by our approach treat the first view's azimuth as the origin.}}  We assume only that the absolute elevation can be sensed using gravity, and that the agent is aware of the relative motion from the previous view. Let $\bm{p}_t$ denote this proprioceptive metadata \JD{(elevation, relative motion)}.

The \textsc{sense} module processes these inputs in separate neural network stacks to produce two vector outputs, which we jointly denote as $\bm{s}_t=\textsc{sense}(\bm{x}_t,\bm{p}_t)$ ({see Fig~\ref{fig:architecture}, top left}). \textsc{fuse} combines information from both input streams and embeds it into $\bm{f}_t=\textsc{fuse}(\bm{s}_t)$ (Fig~\ref{fig:architecture}, top center). Then this combined sensory information $\bm{f}_t$ from the current observation is fed into \textsc{aggregate}, which is a long short term memory module (LSTM)~\cite{hochreiter1997long}. \textsc{aggregate} maintains an encoded internal model $\bm{a}_t$ of the object/scene under observation to ``remember'' all relevant information from past observations. At each timestep, it updates this code, combining it with the current observation to produce $\bm{a}_t=\textsc{aggregate}(\bm{f}_1,\cdots,\bm{f}_t)$ (Fig~\ref{fig:architecture}, top right).

\textsc{sense}, \textsc{fuse}, and \textsc{aggregate} together may be thought of as performing the function of ``encoding'' observations into an internal model. This code $\bm{a}_t$ is now fed into two modules, for producing the output viewgrid and selecting the action, respectively.

\vspace*{-0.1in}
\paragraph{Decoding to the inferred viewgrid} \textsc{decode} translates the aggregated code into the predicted viewgrid $\hat{V}_t(\bm{x}_1,\cdots,\bm{x}_t)=\textsc{decode}(\bm{a}_t)$. To do this, it first reshapes $\bm{a}_t$ into a sequence of small 2D feature maps (Fig~\ref{fig:architecture}, bottom right), before upsampling to the target dimensions using a series of learned up-convolutions.  The final up-convolution produces $MN$ maps, one for each of the $MN$ views in the viewgrid. For color images, we produce $3MN$ maps, one for each color channel of each view. This is then reshaped into the target viewgrid (Fig~\ref{fig:architecture}, bottom center). \JD{Seen views are pasted directly from memory to the appropriate viewgrid positions.} 


\vspace*{-0.1in}
\paragraph{Acting to select the next viewpoint to observe} Finally, \textsc{act} processes the aggregate code $\bm{a}_t$ to issue a motor command $\bm{\delta}_t=\textsc{act}(\bm{a}_t)$ (Fig~\ref{fig:architecture}, middle right). {For objects, the motor commands rotate the object (i.e., agent manipulates the object or peers around it); for scenes, the motor commands move the camera (i.e., agent turns in the 3D environment).}  Upon execution, the {observation's pose} updates for the next timestep to $\bm{\theta}_{t+1}=\bm{\theta}_t + \bm{\delta}_t$. For $t=1$, $\bm{\theta}_1$ is randomly sampled.

Internally, $\textsc{act}$ first produces a distribution over all possible actions, and then samples $\bm{\delta}_t$ from this distribution. Motions in the real world are constrained to be continuous, so we restrict \textsc{act} to select ``small" actions (\JDtwo{details in Sec~\ref{sec:exp}}). Due to the sampling operation, $\textsc{act}$ is a \emph{stochastic} neural network~\cite{neal1990learning}.  Once the new viewpoint $\bm{\theta}_{t+1}$ is set, a new view is captured and the whole process repeats. This happens until $T$ timesteps have passed, involving $T-1$ actions.


\subsection{Objective function and model optimization}\label{sec:objective}

All modules are jointly optimized end-to-end to improve the final reconstructed viewgrid $\hat{V}_T$, which contains predicted views $\bm{\hat{x}_T}(X,\bm{\theta}_j)$ for all viewpoints $\bm{\theta}_j, 1\leq j \leq MN$.


A simple objective would be to minimize the distance between predicted and target views at the same viewpoint coordinate at time $T$: for each {training} object $X$, $L_T(X)=\sum_i d(\bm{\hat{x}_T}(X,\bm{\theta}_i), \bm{x}(X,\bm{\theta}_i))$, where $d(.)$ is a distance function. However, this loss function requires viewpoint coordinates to be registered exactly in the \JDtwo{output} and target viewgrids, \changes{whereas the agent has only partial knowledge of the object's pose (known elevation but unknown azimuth)} and thus must output viewgrids assuming the azimuth coordinate of the first view to be the origin. \JD{Therefore, output viewgrids are shifted by an angle $\Delta_0$ from the target viewgrid, and  $\Delta_0$ must be included in the loss function:}
\begin{equation}
\setlength{\abovedisplayskip}{2pt}
\setlength{\belowdisplayskip}{2pt}
  L_T(X)=\sum_{i=1}^{MN} d(\bm{\hat{x}_T}(X,\bm{\theta}_i+\Delta_0), \bm{x}(X,\bm{\theta}_i)).
  \label{eq:loss}
\end{equation}
We set $d(.)$ to be the per-pixel squared $\mathcal{L}_2$ distance. \JD{With this choice, the agent expresses its uncertainty by averaging over the modes of its beliefs about unseen views. In principle, $d(.)$ could be replaced with other metrics. In particular, a GAN loss~\cite{gan} would force the agent to select one belief mode to produce a photorealistic viewgrid, but the selected mode might not match the ground truth. Rather than one \emph{plausible} photorealistic rendering \KG{(GAN)}, we aim to resolve uncertainty over time to converge to the \emph{correct} model \KG{($\mathcal{L}_2$)}.}

\changes{Note that $\Delta_0$ is used only at training time and only to compute the loss.
\KG{This} choice has the effect of making the setting more realistic and also significantly improving generalization ability. If the viewpoint were fully known, the system might minimize the training objective by memorizing a mapping from $<$view, viewpoint$>$ to viewgrid, which would not generalize. Instead, with our unknown viewpoint setting and training objective (Eq~\ref{eq:loss}), the system is incentivized to learn the harder but more generalizable skill of mental object rotation to produce the target viewgrids.}

To minimize the loss, we employ a combination of stochastic gradient descent (using backpropagation through time to handle recurrence) and REINFORCE~\cite{REINFORCE}, as in~\cite{Mnih2014-dg}. Specifically, the gradient of the loss in Eq~\ref{eq:loss} is backpropagated via the \textsc{decode}, \textsc{aggregate}, \textsc{fuse}, and \textsc{sense} modules. If \textsc{act} were a standard deterministic neural network module, it could receive gradients from \textsc{sense}. However, \textsc{act} is stochastic as it involves a sampling operation. To handle this, we use the REINFORCE technique: we compute reward $R(X)=-L_T(X)$, and apply it to the outputs of \textsc{act} at all timesteps\footnote{\JDtwo{In practice, we reduce the variance of $R$ for stable gradients by subtracting the ``baseline'' expected reward over the last few iterations.}}, backpropagating to encourage \textsc{act} behaviors that led to high rewards. To backpropagate through time (BPTT) to the previous timestep, the reward gradient from \textsc{act} is now passed to \textsc{aggregate} for the previous timestep. BPTT for the LSTM module inside \textsc{aggregate} proceeds normally with incoming gradients from the various timesteps{---namely,} the $\textsc{decode}$ loss gradient for $t=T$, and the \textsc{act} reward gradients for previous timesteps.

In practice, we find it beneficial to penalize errors in the predicted viewgrid at \emph{every} timestep, rather than only at $t=T$, so that the loss $L_T(X)$ of Eq~\ref{eq:loss} changes to: 
\begin{equation}
\setlength{\abovedisplayskip}{2pt}
\setlength{\belowdisplayskip}{2pt}
  L(X)= \sum_{t=1}^T \sum_{i=1}^{MN} d(\bm{\hat{x}_t}(X,\bm{\theta}_i+\Delta_0), \bm{x}(X,\bm{\theta}_i)).
  \label{eq:pertime_loss}
\end{equation}
{Note that this loss $L(X)$ would reduce to the loss $L_T(X)$ of Eq~\ref{eq:loss} if, instead of the summation over $t$, $t$ were held fixed at $T$.} 
Since there are now incoming loss gradients to \textsc{decode} at every timestep, BPTT involves adding reward gradients from \textsc{act} to per-timestep loss gradients from \textsc{decode} before passing through \textsc{aggregate}. BPTT through \textsc{aggregate} is unaffected.
{Our approach learns a non-myopic policy to best utilize the budget $T$, meaning it can learn behaviors more complex than simply choosing the next most promising observation.}
Accordingly, we retain the reward $R(X)=-L_T(X)$ for REINFORCE updates to \textsc{act}, based only on the final prediction; per-timestep rewards would induce greedy short-term behavior and disincentivize actions that yield gains in the long term, but not immediately.

Further, we find it useful to pretrain the entire network with $T=1$, before training \textsc{aggregate} and \textsc{act} with more timesteps, while other modules are frozen at their pretrained configurations. This helps avoid poor local minima and enables much faster convergence.

There are prior methods that use recurrent neural networks and  REINFORCE to achieve some notion of visual attention~\cite{Mnih2014-dg,dinesh-eccv2016,yeung2016end}. Following the best practice of adopting well-honed architectures in the literature, we retain broadly similar architectural choices to these recent instantiations of neural network policy learning where possible.  This also facilitates fair comparisons with~\cite{dinesh-eccv2016} for testing our policy transfer idea (defined below). However, in addition to all the technical details presented above, our approach differs significantly in its objective (see Sec.~\ref{sec:related}).

\subsection{Unsupervised policy transfer to unseen tasks}~\label{sec:policy_transfer_approach}
\vspace*{-0.05in}

The complete scene or 3D object encompasses all potentially useful information for any \KG{task}. 
To capitalize on this property, we next propose an unsupervised policy transfer approach.  The main idea is to inject our generic look-around policy into new unseen tasks in unseen environments.  In particular, we consider transferring our policy---trained without supervision---into a specific recognition task that targets objects unseen by the policy learner.


To do this, we plug in our unsupervised active observation completion policies into the active categorization system of~\cite{dinesh-eccv2016}.  At training time, we train two models: an end-to-end model for active categorization using random policies following~\cite{dinesh-eccv2016} (``model A''), and an active observation completion model (``model B''). Note that our  completion model is, without supervision, trained to look around environments/objects that have zero overlap with model A's target set.  Furthermore, even the \emph{categories} of objects seen during training may differ from those during testing.

At test time, we run forward passes through both models A and B simultaneously. At every timestep, both models observe the same input view. They then communicate as follows: the observation completion model B selects actions to complete its internal model of the new environment. At each timestep, this action is transmitted to model A, in place of the randomly sampled actions that it was trained with. Model A now produces the labels from the correct target label set. If the policy learned in model A is truly generic, it will intelligently explore to solve the new (unseen) categorization task. 


\vspace*{-0.05in}
\section{Experiments}~\label{sec:exp}
\KG{To validate} our approach, we examine the effectiveness of  active completion policies for faster reconstruction (Sec~\ref{sec:active_completion_exp}), as well as their utility for transferring \KG{unsupervised} look-around policies to a recognition task (Sec~\ref{sec:active_completion_exp}).

\subsection{Datasets and experimental setups}~\label{sec:lookaround_exp_setup}
For benchmarking and reproducibility, we {evaluate} active settings with two widely used datasets:

On \textbf{SUN360}~\cite{sun360}, our limited field-of-view (45\degree) agent attempts to complete an omnidirectional scene. SUN360 has spherical panoramas of diverse categories. We use the 26-category subset used in~\cite{sun360,dinesh-eccv2016}. The viewgrid has 32 $\times$ 32 views from 5 camera elevations (-90,-45,\ldots,90\degree) and 8 azimuths (45,90,\ldots,360\degree). At each timestep, the agent moves within a 3 elevations$\times$5 azimuths neighborhood from the current position. \JD{Balancing task difficulty (harder tasks require more views) and training speed (fewer views is faster) considerations, we set training episode length $T=6$ a priori.}

On \textbf{ModelNet}~\cite{modelnet}, our agent manipulates a 3D object to complete its image-based shape model of the object. ModelNet has two subsets of CAD models: ModelNet-40 (40 categories) and ModelNet-10 (10 category-subset of ModelNet-40). To help test our ability to generalize to previously unseen categories, we train on categories in ModelNet-40 that are not in ModelNet-10. We then test both on new instances from the seen categories, and on the unseen categories from ModelNet-10. The viewgrid has 32x32 views from 7 camera elevations (0,$\pm$30,$\pm$60,$\pm$90) and 12 azimuths (30,60,\ldots,360\degree). Per-timestep motions are allowed within the 5$\times$5 neighboring angles of the current viewing angle. The training episode length is $T=4$.

\begin{figure*}[t]
\addtolength{\tabcolsep}{-4pt}
\begin{adjustwidth}{-1.5cm}{-1.75cm}
  \centering
  \vspace{-0.25in}
  \small{\textbf{SUN360 scene observation completion examples}}\\
  \begin{tabular}{cccccc}
    GT viewgrid & & $t=1$ & $t=2$ & $t=3$ & $t=4$\\
    \includegraphics[width=0.23\textwidth,trim={0 5.25cm 0 0.75cm},clip]{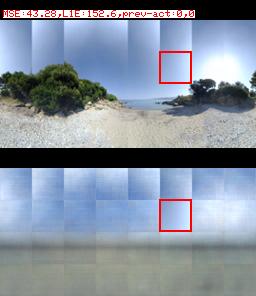} &  &
    \includegraphics[width=0.23\textwidth,trim={0 0.0cm 0 5.95cm},clip]{figs/SUN360_eg/3000570/3000570_SUN360_test0040time001.jpg} &
    \includegraphics[width=0.23\textwidth,trim={0 0.0cm 0 5.95cm},clip]{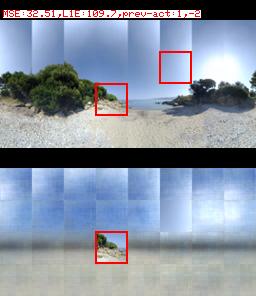} &
    \includegraphics[width=0.23\textwidth,trim={0 0.0cm 0 5.95cm},clip]{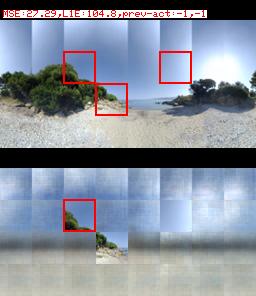} &
    \includegraphics[width=0.23\textwidth,trim={0 0.0cm 0 5.95cm},clip]{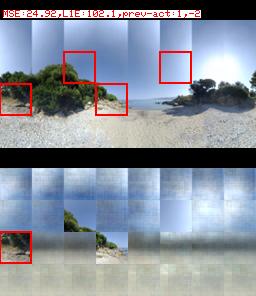}\\
    \includegraphics[width=0.23\textwidth,trim={0 8.65cm 0 0.75cm},clip]{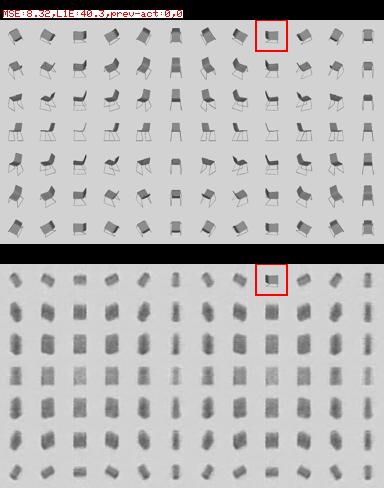} & &
    \includegraphics[width=0.23\textwidth,trim={0 0.0cm 0 9.35cm},clip]{figs/ModelNet30_eg/3000566/3000566_ModelNet30_unseenCat_test0009time001.jpg} &
    \includegraphics[width=0.23\textwidth,trim={0 0.0cm 0 9.35cm},clip]{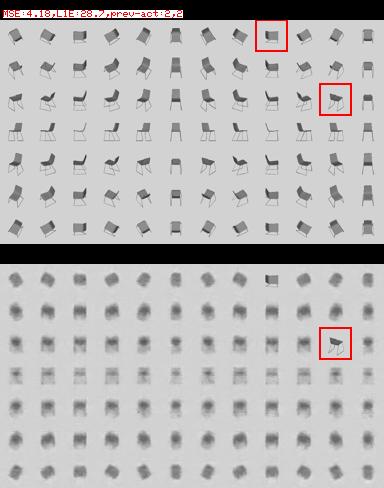} &
    \includegraphics[width=0.23\textwidth,trim={0 0.0cm 0 9.35cm},clip]{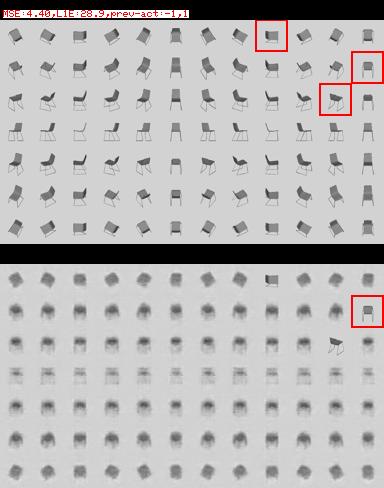} &
    \includegraphics[width=0.23\textwidth,trim={0 0.0cm 0 9.35cm},clip]{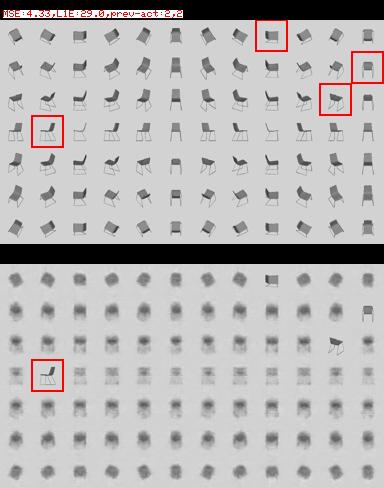}
  \end{tabular}
\end{adjustwidth}
\begin{adjustwidth}{-0.75cm}{-1.0cm}
 \caption{\small{Best viewed on pdf with zoom. \KG{Episodes of active observation completion for a scene (top) and object (bottom).}Column 1 shows the ground truth viewgrid with a red square around the random starting view. Columns 2-5 show our method's viewgrid completions for $t=1,\dots,4$ with red squares around selected views. \JD{As the model's beliefs evolve, the space of possibilities grows more constrained, and the shape of the ground truth viewgrid begins to emerge}. \textbf{Row 1:} The system correctly estimates a flat outdoor scene at $t=1$, inferring the position of a horizon and even the sun from just one view of a gradient in the sky. At $t=2$, it sees rocks and sand, and updates the viewgrid to begin resembling a beach. It then continues to focus on the most interesting (and unpredictable) region of the scene containing the rocks and shrubs. \textbf{Row 2:} The first view is overhead, and azimuthally aligned with one of the sides of an unseen category object (chair). Our agent chooses to move as far from this view as possible at $t=2$, instantly forming a much more chair-like predicted viewgrid, which continues to improve afterwards.}}
  \label{fig:sun360_eg}
\end{adjustwidth}
\vspace{-0.13in}
\end{figure*}

\begin{table*}[t]
  \caption{\small{Per-pixel mean squared error (MSE$\times$1000) with episode length set to training length $T$ (6 on SUN360, 4 on ModelNet), and corresponding improvement over \texttt{1-view} baseline. Lower error and higher improvement is better. RGB (luminance) values in color (gray) images are normalized to [0,1], so error values are on scale of 0 to 1000.}}
  \label{tab:mse}
    \centering
  \resizebox{0.85\textwidth}{!}{
    \begin{tabular}{lcccccc}
      \toprule
      Dataset$\rightarrow$ & \multicolumn{2}{c}{SUN360} & \multicolumn{2}{c}{ModelNet (seen classes)} & \multicolumn{2}{c}{ModelNet (unseen classes)} \\
      \cmidrule(lr){2-3} \cmidrule(lr){4-5} \cmidrule(lr){6-7}
      Method$\downarrow$ | Metric$\rightarrow$ & MSE(x1000)     & Improvement      & MSE(x1000)    & Improvement      & MSE(x1000)    & Improvement    \\
      \midrule
      \texttt{1-view}                                   & 39.40          & -                & 3.83          & -               & 7.38          & -            \\
      \texttt{random}                                   & 31.88          & 19.09\%          & 3.46          & 9.66\%          & 6.22          & 15.72\%          \\
      \texttt{large-action}                             & 30.76          & 21.93\%          & 3.44          & 10.18\%         & 6.16          & 16.53\%          \\
      \texttt{peek-saliency}~\cite{harel2006graph}      & 27.00          & 31.47\%          & 3.47          & 9.40\%          & 6.35          & 13.96\%          \\
      \texttt{ours}                                     & \textbf{23.16} & \textbf{41.22\%} & \textbf{3.25} & \textbf{15.14\%} & \textbf{5.65} & \textbf{23.44\%}         \\
      \bottomrule
    \end{tabular}
  }
\vspace{-0.13in}
\end{table*}

\vspace*{-0.1in}
\paragraph{Baselines} We test our active completion approach ``\texttt{ours}'' against a variety of baselines:
\begin{itemize}
\itemsep-0.5em
  \item \texttt{1-view} is our method trained with $T=1$. No information aggregation or action selection is performed by this baseline.
  \item \texttt{random} is identical to our approach, except that the action selection module is replaced by \KG{randomly} selected actions from the pool of all possible actions.
  \item \texttt{large-action} chooses the largest \changes{allowable} action repeatedly. This tests if ``informative'' views are just far-apart views.
    Since there is no one largest action, we test all actions along the perimeter of the grid of allowable actions, and report results for the best-performing action on the test set.
  \item \texttt{peek-saliency} moves to the most salient view within reach at each timestep, using a popular saliency metric~\cite{harel2006graph}. To avoid getting stuck in a local saliency maximum, it does not revisit seen views. \texttt{peek-saliency} tests if salient views are informative for observation completion. Note that this baseline ``peeks'' at neighboring views prior to action selection to measure saliency, giving it an unfair and impossible advantage over \texttt{ours} and the other baselines.
\end{itemize}
These baselines all use the same network architecture as \texttt{ours}, differing only in the exploration policy which we seek to evaluate.

\begin{figure*}[t]
  \centering
    \vspace{-0.31in}
  \includegraphics[height=0.20\textwidth,trim={0 0.3cm 0cm 0cm},clip]{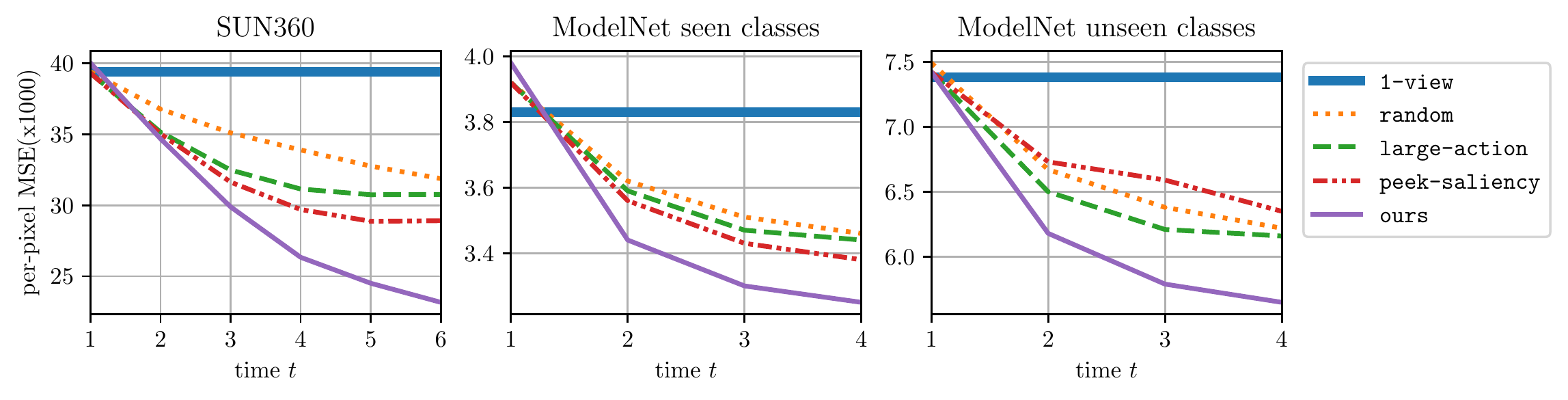}
  \caption{\small{Active observation completion: per-pixel mean-squared error versus time for the three test datasets. 
  }}
  \label{fig:mse_vs_t}
\vspace{-0.13in}
\end{figure*}

\vspace{-0.05in}
\subsection{Active observation completion results}~\label{sec:active_completion_exp}
Tab~\ref{tab:mse} shows the scene and object completion mean-squared error on SUN360 and ModelNet (seen and unseen classes). For these results, episode lengths are held constant to $T$ timesteps \cc{(6 on SUN360, 4 on ModelNet)}, same as during training.
While all the multi-view methods improve over \texttt{1-view}, our method outperforms all baselines by large margins. To isolate the impact of view selection, we report improvement over \texttt{1-view} for all methods. Compared to \texttt{random}, \texttt{ours} consistently yields approximately 2x improvement; {our gains over \texttt{large-action} are also substantial in all cases, meaning that simply looking at well-spaced views is not enough.} {Both outcomes highlight the major value in learning to intelligently look around.} Improvements are larger on more difficult datasets, where errors are larger (SUN360 $>$ ModelNet unseen $>$ ModelNet seen). This is as expected, since additional views are most critical where one view produces very poor results. On SUN360, \texttt{peek-saliency}, which has unfair access to neighboring views for action selection, is the strongest baseline, but still falls short of \texttt{ours}. On ModelNet data, \texttt{peek-saliency} performs poorly, likely because saliency fails to differentiate well between the synthetic CAD model views; what is informative about an object's shape is much more complex than what low-level unsupervised saliency can measure.  Importantly, our advantages hold \emph{even for unseen categories} (rightmost), emphasizing the task-independence of our look-around policies.

Does our approach simply exploit its knowledge of camera elevation to sample useful elevations more than others? For instance, perhaps views from a horizontal camera position (elevation $0\degree$) are more informative than others. Upon investigation, we find that this is not the case in practice. In particular, our learned policy samples all elevations uniformly on both SUN360 and ModelNet data.  Hence, the ability to sense gravity alone offers no advantage over the \texttt{random} baseline.

Figure~\ref{fig:mse_vs_t} further shows how error behaves as a function of time.  With perfect information aggregation, all methods should asymptotically approach zero error at high $t$, which diminishes the value of intelligent exploration.\footnote{This is also a reason to limit the training time budget to $T=4$ \KG{to $6$}.} All methods show consistent improvement, with sharpest error drops for \texttt{ours}.

Fig~\ref{fig:sun360_eg} presents some completion episodes (see \url{https://goo.gl/BgWX3W} for more).
As our system explores, the rough ``shape'' of the target scene or object emerges in its viewgrid predictions.  
We stress that the goal of our work is not to obtain photorealistic images. Rather, the goal is to learn policies for looking around that efficiently resolve model uncertainty in novel environments; the predicted viewgrids visualize the agent's beliefs over time.
The key product of our method is a \emph{policy}, not an image---\KG{as the next result emphasizes.}

\subsection{Unsupervised policy transfer results}~\label{sec:policy_transfer}
\vspace{-0.1in}

Having shown our approach successfully trains unsupervised policies to acquire useful visual observations, we next test how well this policy transfers to a new task with new data from unseen categories (cf.~Sec~\ref{sec:policy_transfer_approach}).


\JD{We closely follow the active categorization experimental setups described in~\cite{dinesh-eccv2016}. 
Using our method presented in Sec~\ref{sec:policy_transfer_approach}, we plug our unsupervised active observation completion policies into the active categorization system of~\cite{dinesh-eccv2016}. The active categorization model (``model A'') is trained with random policies---this is the same as the \texttt{random-policy} baseline below. For ModelNet, we train ``model A'' on ModelNet-10 training objects, and the active observation completion model (``model B'') on \emph{ModelNet-30} training objects, disjoint from the target ModelNet-10 dataset classes. For SUN360, both models are trained on SUN360 training data.}

\vspace*{-0.1in}
\paragraph{Baselines} \JD{As baselines, we consider: 1) \texttt{sup-policy}, the full end-to-end active categorization system trained using the ``Lookahead active RNN'' approach of~\cite{dinesh-eccv2016}; 2) \texttt{1-view}, a passive feed-forward neural network which only processes one randomly presented view and predicts its category. Its architecture is identical to \texttt{sup-policy} minus the action selection  and information aggregation modules; and 3) \texttt{random-policy}, an active categorization system trained on the target classes, which selects random actions. This uses the same core architecture as \texttt{sup-policy}, except for the action selection module. In place of learned actions, it selects random legal motions from the same motion neighborhood as \texttt{sup-policy}.}

\begin{figure}
  \hspace*{-0.4cm}
  \centering
  \includegraphics[width=0.52\textwidth]{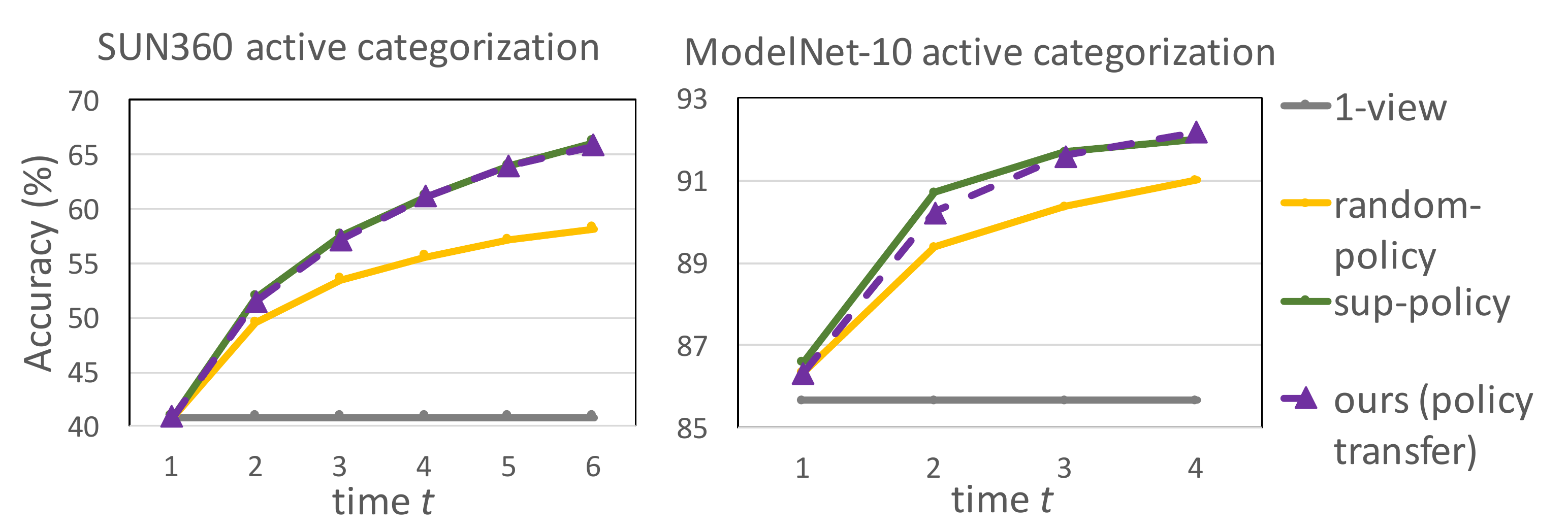}
  \caption[Policy transfer results]{Active categorization accuracy vs.~time on SUN360 scenes (left) and ModelNet-10 objects (right).} 
  \label{fig:policy_transfer}
\vspace{-0.13in}
\end{figure}

\JD{Fig~\ref{fig:policy_transfer} shows the results. For both SUN360 active scene recognition and ModelNet-10 active object recognition, our unsupervised policies perform on par with the end-to-end active categorization policy of~\cite{dinesh-eccv2016}, easily outperforming \texttt{random-policy} and \texttt{1-view}. This is remarkable because the policy being employed in \texttt{ours (policy transfer)} is only trained for the separate, unsupervised active observation completion task. \KG{Further, in the ModelNet case, it is also trained on data from disjoint classes.}}

\JD{This result shows the potential of unsupervised exploratory tasks to facilitate policy learning on massive unlabeled datasets. Policy learning is famously expensive in terms of data, computation, and time.  Once trained, exploratory policies like the proposed active completion framework could be transferred to arbitrary new tasks with much smaller datasets. Performance may further improve if instead of directly transferring the policy, the policy could be finetuned for the new task, analogous to feature finetuning as widely employed in the passive recognition setting.}

\vspace{-0.1in}
\section{Conclusions}~\label{sec:conc}
Our work tackles a new problem: how can a visual agent learn to look around, independent of a recognition task? We presented a new active observation completion framework for general exploratory behavior learning. Our reinforcement learning solution demonstrates consistently strong results across very different settings for realistic scene and object completion, compared to multiple \KG{revealing} baselines. Our results showing successful application of our unsupervised exploratory policy for active recognition are the first demonstration of ``policy transfer'' between tasks to our knowledge. These results hold great promise for 
\KG{task-agnostic} exploration, an important step towards autonomous embodied visual agents.


%
%
\pagebreak
{\small
\bibliographystyle{ieee}
\bibliography{jd_refs}
}

\end{document}